\documentclass[conference, a4paper]{IEEEtran}

\usepackage{comment}
\usepackage{graphicx}
\usepackage{multirow}
\usepackage{cite}
\usepackage{longtable}
\usepackage{xurl}

\begin{document}

\title{White blood cell subtype detection and classification}

\author{\IEEEauthorblockN{Nalla Praveen}
\IEEEauthorblockA{\textit{Dept. of Information Technology} \\
\textit{Indian Institute of Information Technology Allahabad}\\
Prayagraj, India \\
mit2019038@iiita.ac.in} \\
\IEEEauthorblockN{Sanjay Kumar Sonbhadra}
\IEEEauthorblockA{\textit{Dept. of Information Technology} \\
\textit{Indian Institute of Information Technology Allahabad}\\
Prayagraj, India \\
rsi2017502@iiita.ac.in} \\
\IEEEauthorblockN{M. Syafrullah}
\IEEEauthorblockA{\textit{Program of Master of Computer Science} \\
\textit{Universitas Budi Luhur}\\
Indonesia \\
mohammad.syafrullah@budiluhur.ac.id}
\and
\IEEEauthorblockN{Narinder Singh Punn}
\IEEEauthorblockA{\textit{Dept. of Information Technology} \\
\textit{Indian Institute of Information Technology Allahabad}\\
Prayagraj, India \\
pse2017002@iiita.ac.in} \\
\IEEEauthorblockN{Sonali Agarwal}
\IEEEauthorblockA{\textit{Dept. of Information Technology} \\
\textit{Indian Institute of Information Technology Allahabad}\\
Prayagraj, India \\
sonali@iiita.ac.in}\\
\IEEEauthorblockN{Krisna Adiyarta}
\IEEEauthorblockA{\textit{Program of Master of Computer Science} \\
\textit{Universitas Budi Luhur}\\
Indonesia \\
krisna.adiyarta@gmail.com}
}

\maketitle

\begin{abstract}
Machine learning has endless applications in the health care industry. White blood cell classification is one of the interesting and promising area of research. The classification of the white blood cells plays an important part in the medical diagnosis. In practise white blood cell classification is performed by the haematologist by taking a small smear of blood and careful examination under the microscope. The current procedures to identify the white blood cell subtype is more time taking and error-prone. The computer aided detection and diagnosis of the white blood cells tend to avoid the human error and reduce the time taken to classify the white blood cells. In the recent years several deep learning approaches have been developed in the context of classification of the white blood cells that are able to identify but are unable to localize the positions of white blood cells in the blood cell image. Following this, the present research proposes to utilize YOLOv3 object detection technique to localize and classify the white blood cells with bounding boxes. With exhaustive experimental analysis, the proposed work is found to detect the white blood cell with 99.2\% accuracy and classify with 90\% accuracy.
\end{abstract}


\IEEEpeerreviewmaketitle

\section{Introduction}
\begin{figure}
\centering
\includegraphics[width=\linewidth]{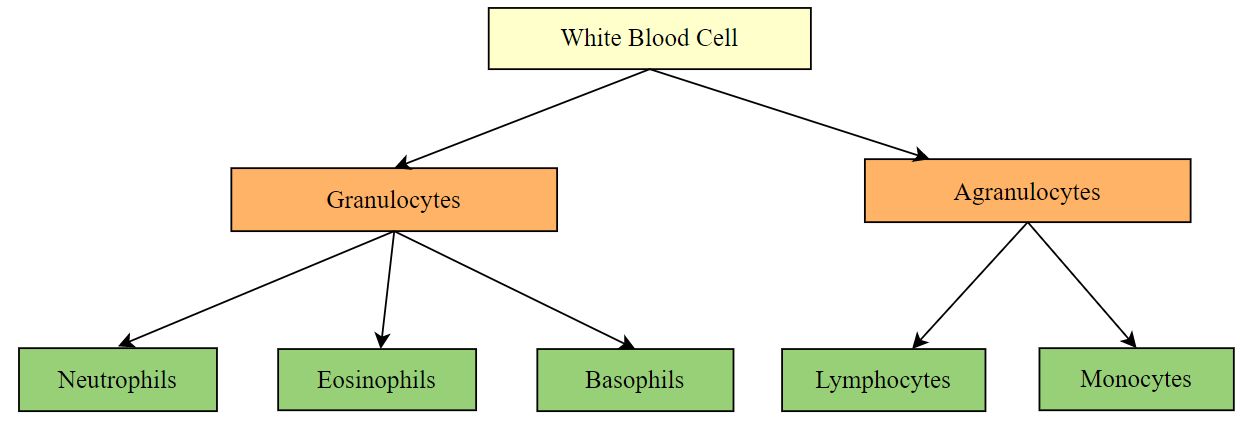}
\caption{White blood cell sub types classification.} \label{fig1}
\end{figure}

Human blood is composed of the blood cells which accounts for 45\% of the blood tissue by volume, whereas the remaining 55\% of the volume composed of plasma (the liquid portion of the blood). There are three types of blood cells: red blood cells (erythrocytes), white blood cells (leukocytes), platelets (thrombocytes). White blood cells are produced in the lymphoid tissue and bone marrow. These cells are classified into two types – granulocytes and agranulocytes. Granulocytes are further classified into neutrophils, basophils and eosinophils, whereas agranulocytes are classified into monocytes and lymphocytes, as shown in Fig.~\ref{fig1}. Each type of cells play a different role to fight against the infections or diseases~\cite{WBC}. Neutrophils helps in healing the damaged tissues and resolve infections. Unusual levels of basophils result in severe types of blood disorders. Eosinophils are disease fighting white blood cells. Whereas, lymphocytes fight against bacteria, virus and the cells that threaten its functioning; monocytes fight against infections, remove dead or damaged tissues, destroy cancer cells. Each of these WBCs differ in the shape of nucleus, color and texture, thereby making it a challenging task for manual analysis.

White blood cell count is done by the haematologist manually. Several procedures are required to determine the location and classify the type of the white blood cells. The current techniques to classify the white blood cells are error prone and time consuming. To reduce the time taken for the blood cell classification and reduce the human errors the computer aided techniques are utilized to automatically classify and detect the location of the white blood cells in the blood cell image. The image classification models such as CNN~\cite{article21,inproceedings4}, VGG-16~\cite{article24}, ResNet~\cite{inproceedings5} are efficient and are widely used for various applications and domain. The previous works proposed on the white blood cells mainly focused on image segmentation~\cite{unknown} and classification.

The rest of the paper is structured as follows- Section II presents the related work in the field of research, Section III presents the object detection technique, YOLOv3~\cite{article21,inproceedings7}, Section IV presents the proposed methodology, section V presents the results, section VI presents the conclusion of the present research work.

\section{Related Work}
In the recent years, several deep learning methods have been proposed to classify the white blood cells. Diouf et al.~\cite{article5} proposed a seven layered convolution neural network that uses standard convolution and max pooling operations to distinguish the four types of white blood cells. The proposed model was able to classify with 97\% accuracy. Mu-ChunSu et al.~\cite{article3} proposed a new segmentation algorithm for segmenting a white blood cell from the smear image. The proposed method involves three stages: (1) segmentation of WBC, (2) extraction of features, (3) design of a classifier. The segmentation algorithm finds the region of WBC on the HSI color space. In the second stage, three kinds of features are extracted for classification: geometrical, color and local directional pattern (LDP) features. Following this, three kinds of neural network classifiers are used: multi layer perceptron~\cite{article20}, support vector machine and hyper rectangular composite neural networks for classification that achieved 99.1\%, 97.5\% and 88.9\% accuracy respectively.  Ozyurt~\cite{article13} proposed a fused CNN model for WBC detection with MRMR feature selection and extreme learning machine. This approach is able to classify WBC with an accuracy of 96\%. Vatathanavaro et al.~\cite {article12} explained the study of two CNN architectures: VGG-16 and ResNet-50 to classify five types of white blood cells. ResNet-50 is the best classifer and it achieved 88.3\% accuracy. Theera-Umpon~\cite {article17} utilized a new set of features generated using the mathematical morphology to extract nucleus features. Later, these features were used to train the artificial neural network and Bayes classifer to identify the type of WBC. The authors achieved promising and robust results by adapting the five fold cross validation on blood cell images. 

Recently, Almezhghwi et al.~\cite{article4} proposed a generative adversarial network for image augmentation and state-of-the-art deep neural network for classifying the white blood cells. The proposed method classifies the white blood cells with an accuracy of 98.8\%. Madhumala et al.~\cite{inproceedings} proposed a systematic approach to classify the white blood cells through statistical pattern analysis. A Naive Bayes approach is used for the classification of the white blood cells depending on the roundness, formfactor, solidity and compactness shape features. The overall accuracy obtained using this approach is 83.2\%. Habibzadeh et al. \cite{article8} proposed a method to classify and divide the white blood cells using K-means clustering, whereas SVM~\cite{article14} and neural network were used for classification. The authors achieved 80\% accuracy in classifying white blood cells. Liqun et al.~\cite{article27} improved classification accuracy based on the feature weight K-means clustering for extracting white blood cells. 

However, it can be observed that the standalone classification of the WBCs is not sufficient for further analysis. Following this context, the present article proposes YOLOv3 based localization and classification of WBCs using blood cell images.

\section{Materials and methods} 
\subsection{Dataset}
The dataset used is publicly available Blood Cell Count Detection (BCCD) dataset~\cite{BCI}. BCCD dataset consists of 364 images with annotations and labels of lymphocyte, monocyte, neutrophil and eosinophil cells. The images are RGB of size 320$\times$240$\times$3 and are in jpeg format. The data samples are increased with the image augmentation techniques such as rotation, flip, noise, etc. BCCD is MIT licensed dataset available in Kaggle. There are 2497 Eosinophil images, 2483 Lymphocyte images, 2487 Monocyte images and 2499 Neutrophil images in the training dataset. There are 574 Eosinophil images, 620 Lymphocyte images, 620 Monocyte images and 616 Neutrophil images in the testing dataset.

\subsection{YOLOv3}
YOLO~\cite{inproceedings6} is shortened form of you only look once and it uses convolutional neural network for object detection. YOLO can detect multiple objects on a single image. Apart from predicting classes of the objects, YOLO also detects the locations of the objects on the image. YOLO applies a single neural network to the whole image that divides image into several regions and produces the probabilities for every region. YOLO predicts the number of bounding boxes that cover some regions on the image and chooses the best ones according to the probabilities.
\begin{figure}
\centering
\includegraphics[width=\linewidth]{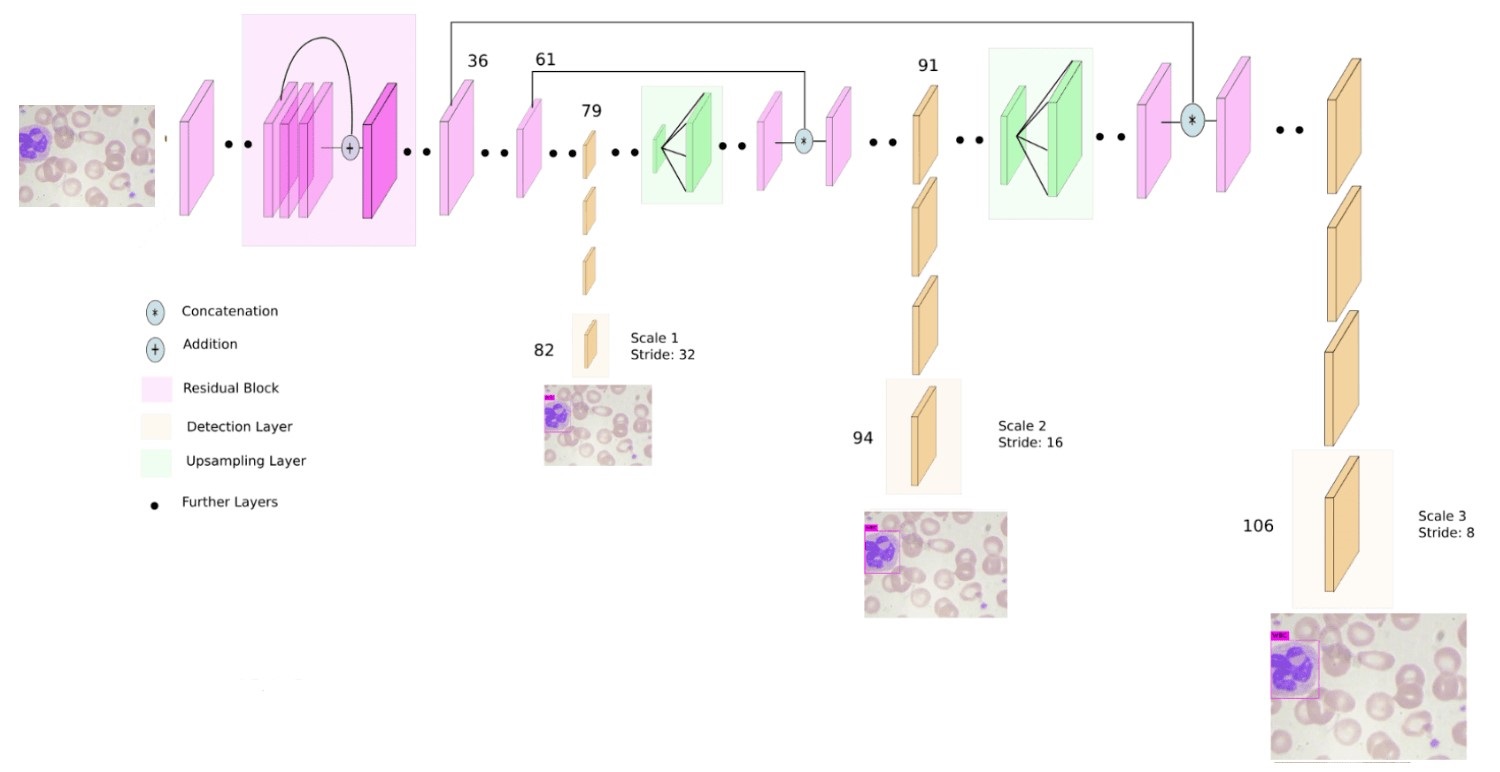}
\caption{Schematic representation of the YOLOv3 network architecture.} \label{fig2}
\end{figure}

\begin{figure}
\centering
\includegraphics[width=\linewidth]{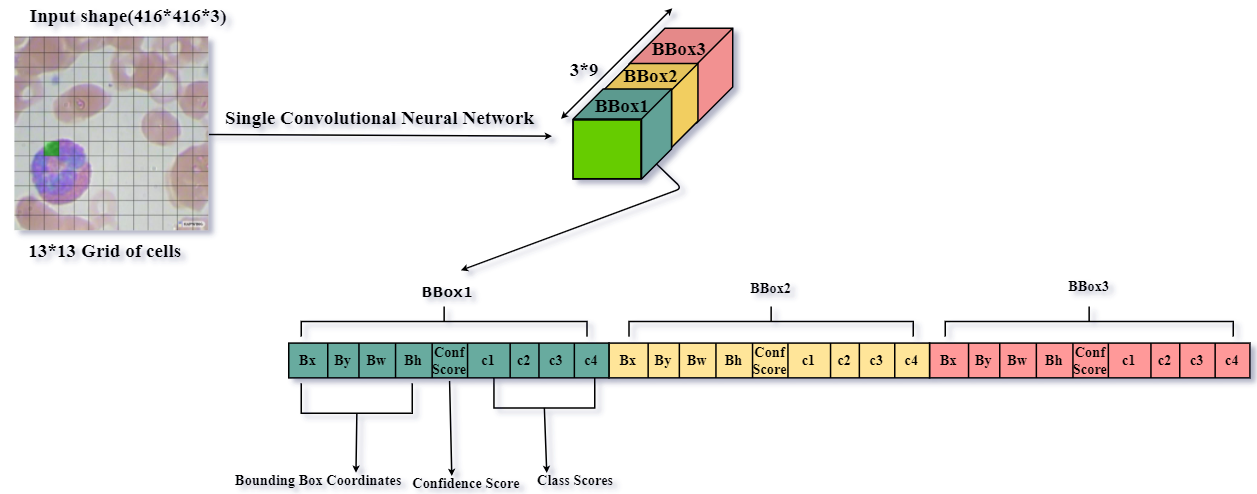}
\caption{YOLOv3: detection and classification work flow diagram.} \label{fig3}
\end{figure}

YOLOv3 consists of 53 convolutional layers that are called as Darknet-u53. For the detection task 53 more layers are added that gives the 106 layered architecture for YOLOv3. The detections are made at 82, 94, 106 layers as shown in Fig.~\ref{fig2}, where the network downsamples the feature map by using strides of 32, 16 and 8. To produce the desired output, YOLOv3 applies 1$\times$1 detection kernels at these layers in the network. The shape of the detection kernels has its depth that is calculated by the expression $(b\times(5+c))$, where b is the number of bounding boxes the YOLOv3 predicts i.e 3 for every cell of the feature maps and $c$ is the number of classes. Each bounding box has $(5+c)$ attributes such as width, height, center coordinates of the bounding box, objectness score, list of confidence score for every class that this cell in the bounding box might belong, as shown in Fig.~\ref{fig3}. There are no pooling layers in YOLOv3, instead convolutional layers are used to down sample the feature maps and prevent the loss of the low-level features that tend to improve the detection of the smaller objects.

In order to predict the bounding boxes, YOLOv3 uses the predefined bounding boxes that are called anchors. These anchors are used to predict the real width and real height of the bounding box. Nine anchor boxes are used i.e three anchor boxes for each scale. To calculate the anchors k-means clustering is used. Later, YOLOv3 calculates the offsets to these anchors and better fit the region of interest with the bounding boxes and finally to compute the center coordinates of the bounding boxes, YOLOv3 passes the outputs through the sigmoid function. The below equations are used to find the width, height and center coordinates of the bounding boxes:
\begin{equation}
\mathcal{B}_{x}\,=\,\sigma(\mathcal{T}_{x})\,+\,\mathcal{C}_{x}
\label{eq_1}
\end{equation}
\begin{equation}
\mathcal{B}_{y}\,=\,\sigma(\mathcal{T}_{y})\,+\,\mathcal{C}_{y}
\label{eq_2}
\end{equation}
\begin{equation}
\mathcal{B}_{w}\,=\,(\mathcal{P}_{w})\,*\,exp(\mathcal{T}_{w})
\label{eq_3}
\end{equation}
\begin{equation}
\mathcal{B}_{h}\,=\,(\mathcal{P}_{h})\,*\,exp(\mathcal{T}_{h})
\label{eq_4}
\end{equation}
where, $\mathcal{B}_{x}, \mathcal{B}_{y}$ are the center coordinates of the bounding box; $\mathcal{B}_{h}, \mathcal{B}_{w}$ is the height and the width of the bounding box respectively; $\mathcal{T}_{x}, \mathcal{T}_{y}, \mathcal{T}_{w}, \mathcal{T}_{h}$ are the output of the neural network.; $\mathcal{C}_{x}, \mathcal{C}_{y}$ are the coordinates of the top left corner of the cell on the grid of the anchor box; $\mathcal{P}_{w}, \mathcal{P}_{h}$ are the width and height of anchor.

\section{Proposed method}
The proposed WBC classification and detection method is based on the YOLOv3 object detection algorithm. The proposed method mainly consists of two phases. Phase 1 is WBC bounding box detection from the blood cell image, Phase 2 is classification and detection of WBC subtypes.

\subsection{Phase 1 - WBC bounding box generation}
\begin{figure}
\centering
\includegraphics[width=\linewidth]{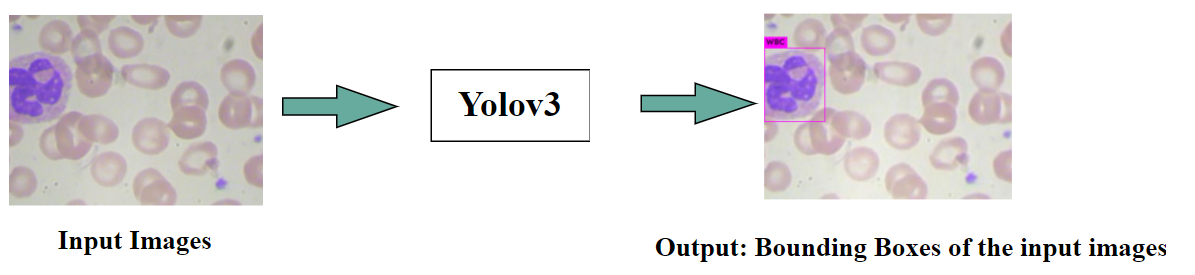}
\caption{Phase 1: bounding box generation for detection of WBC work flow diagram.} \label{fig4}
\end{figure}

Phase 1 generates the bounding boxes that serves as the training set for phase 2. In phase 1 364 white blood cell images and their corresponding annotations are used for training. The images are preprocessed along with the annotation file that consists of bounding box coordinates to extract the white blood cells. YOLOv3 uses these images and coordinates to generate the bounding boxes around white blood cells. These generated bounding boxes along with confidence score are considered as training set for phase 2. This phase can also be used to count RBC, WBC, platelets from the blood cell images~\cite{second,first}. This phase detects the bounding box of a single class (WBC) as shown in Fig.~\ref{fig4}. The shape of the detection kernel is 3$\times$6 i.e calculated by the expression $(b*(5+c))$, where $b=3$ and $c=1$ (for single class detection).

\subsection{Phase 2- Classification and detection of WBC subtypes}
It is a multi-class (four classes) classification problem. The input to the phase 2 is the training dataset images and the bounding box of the respective images that are generated by the output of phase 1. The output of this phase is the class predicted, bounding box of the subtype of white blood cell in the blood cell image, class probabilities and confidence score of the bounding box.

\begin{figure}
\centering
\includegraphics[width=\linewidth]{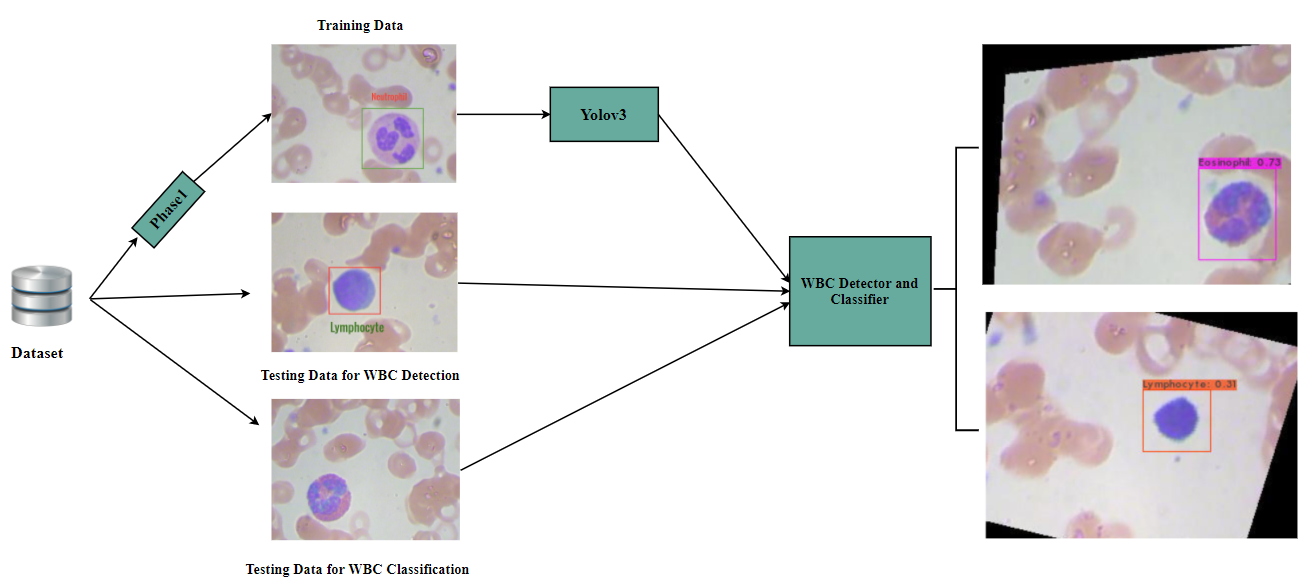}
\caption{Overall schematic representation of the proposed framework for detection and classification of the WBC.} \label{fig5}
\end{figure}

The shape of the detection kernel is $3\times9$ is calculated by the expression $(b\times(5+c))$, where $b=3$ and $c=4$ (four class classification). The overall workflow diagram is shown in Fig.~\ref{fig5}.

\subsection{Evaluation with test data}
For classification the considered evaluations metrics are accuracy, precision, recall, f1-Score. The entire test dataset is used for evaluation of classification task i.e 2430 images of four different classes. These metrics are calculated by the following equations:
\begin{equation}
Accuracy \,=\,\frac{TP+TN}{TP+TN+FP+FN}
\label{eq_5}
\end{equation}
\begin{equation}
Precision \,=\,\frac{TP}{TP+FP}
\label{eq_6}
\end{equation}
\begin{equation}
Recall \,=\,\frac{TP}{TP+FN}
\label{eq_7}
\end{equation}
\begin{equation}
F1\mathrm{-Score} \,=\frac{2*Recall*Precision}{Recall+Precision}
\label{eq_8}
\end{equation}

The same metrics are used for evaluating the detection task. The part of the dataset is used for testing the detection. For this, the 200 samples are randomly selected with equal distribution of classes, manually draw the bounding boxes representing WBC region in the images. The detection is true positive if predicted class matches the actual class, bounding box confidence is greater than 0.20 and the predicted bounding box intersection over union (IoU) score is greater than 0.40. The detection is false positive if predicted class does not match the actual class and the predicted bounding box IoU score is less than 0.40. The detection is considered false negative if the detected bounding box confidence is less than 0.20.

\begin{table}[!b]
\begin{center}
\caption{Comparison of classification results over two models.}\label{tab2}
\begin{tabular}{ |c|c|c|c|c|c| } 
\hline
Classifier & WBC Type & F1-Score & Precision & Recall \\
\hline
\multirow{4}{5em}{YOLOv3} & Eosinophil & 0.93 & 0.90 & 0.97 \\ 
& Lymphocyte & 1.0 & 1.0 & 1.0 \\ 
& Monocyte & 0.85 & 0.99 & 0.74 \\ 
& Neutrophil & 0.82 & 0.76  & 0.90 \\
\hline
\multirow{4}{5em}{Faster RCNN + VGG 16} & Eosinophil & 0.93 & 0.88 & 0.97 \\
& Lymphocyte & 0.99 & 0.98 & 1.0 \\ 
& Monocyte & 0.85 & 0.97 & 0.75 \\ 
& Neutrophil & 0.81 & 0.77  & 0.87 \\

\hline
\end{tabular}
\end{center}
\end{table}

\begin{table}
\begin{center}
\caption{Comparison of detection results over two models.}\label{tab3}
\begin{tabular}{ |c|c|c|c|c|c| } 
\hline
Detection & WBC Type & F1-Score & Precision & Recall \\
\hline
\multirow{4}{5em}{YOLOv3} & Eosinophil & 1.0 & 1.0 & 1.0 \\ 
& Lymphocyte & 1.0 & 1.0 & 1.0 \\ 
& Monocyte & 1.0 & 1.0 & 1.0 \\ 
& Neutrophil & 0.97 & 0.97  &   1.0 \\
\hline
\multirow{4}{5em}{Faster RCNN + VGG 16} & Eosinophil & 0.97 & 0.96 & 1.0 \\
& Lymphocyte & 1.0 & 1.0 & 1.0 \\ 
& Monocyte & 1.0 & 1.0 & 1.0 \\ 
& Neutrophil & 0.90 & 0.87  & 1.0 \\

\hline
\end{tabular}
\end{center}
\end{table}

\section{Results and discussion}
The proposed model is able to classify white blood cells subtype with 90\% accuracy. The experimental results of YOLOv3 model compared to faster RCNN using VGG16 for classification and detection tasks are shown in table~\ref{tab2} and table~\ref{tab3}. The YOLOv3 was able to achieve better results due to its ability to consider the contextual information about the available classes by analysing the entire image in a single go, thereby also making it faster than any other model. This also enables the network to effectively draw the localization and identification of relatively smaller object. 

\section{Conclusion}
In this paper, we discussed about detection or localization and classification of white blood cell (WBC) subtypes using blood cell images. The proposed approach utilizes YOLOv3 model divided into two phases, where in phase 1 WBC bounding box is generated and in phase 2 WBC subtypes are classified and detected. The phase 1 approach can also be extended for counting the number of white blood cells, red blood cells and platelets from blood cell images. The proposed approach outperforms the faster RCNN based VGG16 model in both detection and classificaiton of WBC subtypes. The research can be further extended trying different YOLO versions, fine tuning the model and augmenting the dataset. As an extension to this proposed work more object detection and classification techniques can be explored to achieve better results.

\section*{Acknowledgment}
	We thank our institute, Indian Institute of Information Technology Allahabad (IIITA), India and Big Data Analytics (BDA) lab for allocating the centralised computing facility and other necessary resources to perform this research. We extend our thanks to our colleagues for their valuable guidance and suggestions.

\bibliographystyle{IEEEtran}
\bibliography{sample}

\end{document}